\pdfoutput=1

\documentclass[11pt]{article}

\usepackage[]{acl}

\usepackage{times}
\usepackage{latexsym}

\usepackage[T1]{fontenc}

\usepackage[utf8]{inputenc}

\usepackage{microtype}

\usepackage{hyperref}
\usepackage{url}
\usepackage{amsmath,amsthm,amssymb}
\usepackage{enumitem}
\usepackage{algorithmic}
\usepackage{mathtools}
\usepackage{multirow}
\usepackage{booktabs}
\usepackage{graphicx}
\usepackage{caption}
\usepackage{subcaption}
\usepackage{float}
\usepackage{dsfont}
\usepackage{array}
\usepackage{wrapfig}
\usepackage[ruled, lined, linesnumbered, commentsnumbered, longend]{algorithm2e}

\newcommand{\cref}[1]{\S\ref{#1}}

\newcommand{\dataset}{\textsc{Implication}}
\newcommand{\random}{\textsc{Random}}
\newcommand{\dataseta}{ReClor}
\newcommand{\datasetb}{LogiQA}

\newcommand{\method}{\textsc{Apollo}}
\newcommand{\selective}{\textsc{s-MLM}}
\newcommand{\classification}{\textsc{e-CLS}}
\newcommand{\baselinea}{LRReasoner}
\newcommand{\baselineb}{\textsc{Focal Reasoner}}
\newcommand{\baselinec}{MERIt}

\title{\method{}: A Simple Approach for \underline{A}daptive \underline{P}retraining \underline{o}f \underline{L}anguage Models for \underline{Lo}gical Reasoning}

\author{ 
\textbf{Soumya Sanyal}\textsuperscript{{$1$}}\thanks{~~Work done during internship at Microsoft Cognitive Service Research group}\quad
\textbf{Yichong Xu}\textsuperscript{{$2$}}\quad
\textbf{Shuohang Wang}\textsuperscript{{$2$}}\quad
\textbf{Ziyi Yang}\textsuperscript{{$2$}}\\
\textbf{Reid Pryzant}\textsuperscript{{$2$}}\quad
\textbf{Wenhao Yu}\textsuperscript{{$3$}}\footnotemark[1]\quad
\textbf{Chenguang Zhu}\textsuperscript{{$2$}}\quad
\textbf{Xiang Ren}\textsuperscript{{$1$}} \\
{\textsuperscript{$1$}University of Southern California} {\textsuperscript{$2$}Microsoft Cognitive Service Research}\\ {\textsuperscript{$3$}University of Notre Dame} \\
{\texttt{soumyasa@usc.edu}}
}

\begin{document}
\maketitle

\begin{abstract}
Logical reasoning over text is an important ability that requires understanding the semantics of the text and reasoning through them to arrive at correct inferences.
Prior works on pretraining language models to improve the logical reasoning ability require complex processing of training data (e.g., aligning symbolic knowledge to text), yielding task-specific solutions that are not easy to adapt to any general text corpus.
In this work, we propose \method{}, a simple adaptive pretraining approach to improve the logical reasoning skills of language models.
We select a subset of Wikipedia for adaptive pretraining using a set of logical inference keywords as filter words. Further, we propose two self-supervised loss functions for training. First, we modify the masked language modeling loss to mask specific parts-of-speech words that likely require higher-order reasoning to predict them. Second, we propose a sentence-level classification loss that teaches the model to distinguish between entailment and contradiction types of sentences. The proposed pretraining paradigm is both simple and independent of task formats. We demonstrate the effectiveness of \method{} by comparing it with prior baselines on two logical reasoning datasets. \method{} performs comparably on \dataseta{} and outperforms baselines on \datasetb{}. The code base has been made publicly available.\footnote{\url{https://github.com/INK-USC/APOLLO}}
\end{abstract}
\begin{figure}[t]
	\centering
	\includegraphics[width=\columnwidth]{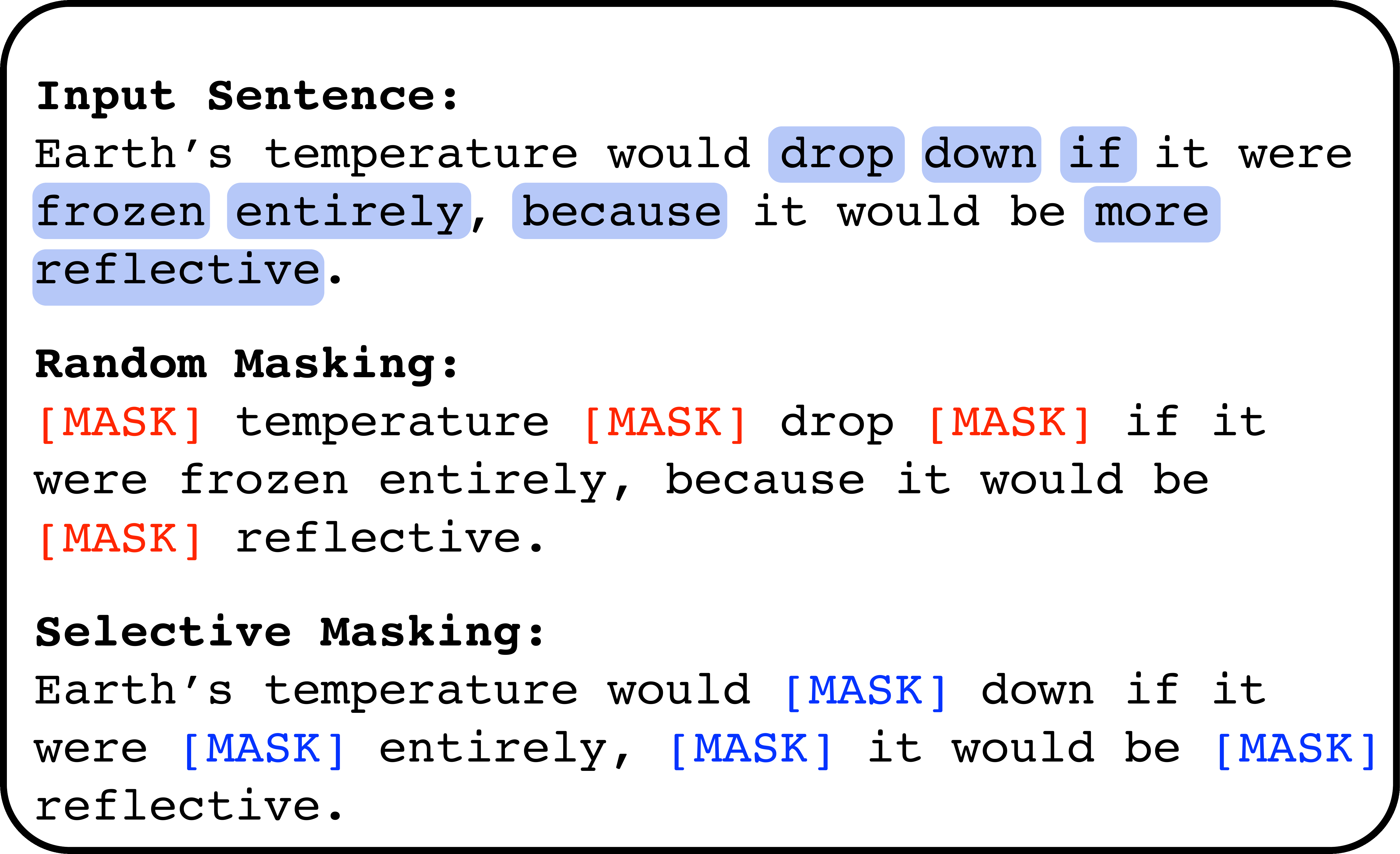}
	\caption{\small \textbf{Motivation of Selective Masking.} In random masking \cite{devlin2019bert}, a word is masked at random. Predicting these words often require more of language understanding than higher-order reasoning (e.g., predicting ``would'' at the $2^{nd}$ {\color{red} [MASK]} place). In selective masking, a word is masked if its POS tag is from a specific set. These candidate words are marked in the blue box in the input sentence. Filling these words requires more reasoning (e.g., to predict ``more'' at the $2^{nd}$ {\color{blue} [MASK]} place instead of ``less'', which is also grammatically valid, the model needs a better understanding of the semantics of the sentence).
	}
	\label{fig:example}
	\vspace{-0.5cm}
\end{figure}

\section{Introduction}
Logical reasoning is an important ability of humans that helps us in making rational decisions based on known information. It is an important ability for text understanding across various downstream tasks, e.g., in open-domain question answering \citep{yang-etal-2018-hotpotqa,zhu2021retrieving}, machine reading comprehension (MRC) \citep{mrc_survey}, etc. Recently, there has been an increasing focus on evaluating the logical reasoning abilities of language models by using MRC tasks that specifically require a significant amount of logical reasoning to obtain the correct answer \citep{reclor_dataset,logiqa_dataset}. In these datasets, the model needs to understand a given context, reason logically about a question to infer new conclusions, and then select the correct answer from a set of options. With the advent of large pre-trained language models (PLMs) in NLP \citep{devlin2019bert,radford2019language,raffel2019exploring}, understanding and improving the logical reasoning abilities of these models has become even more important as these are increasingly being used across a wide variety of real-world tasks.

\begin{figure*}[t]
	\centering
	\includegraphics[width=0.9\textwidth]{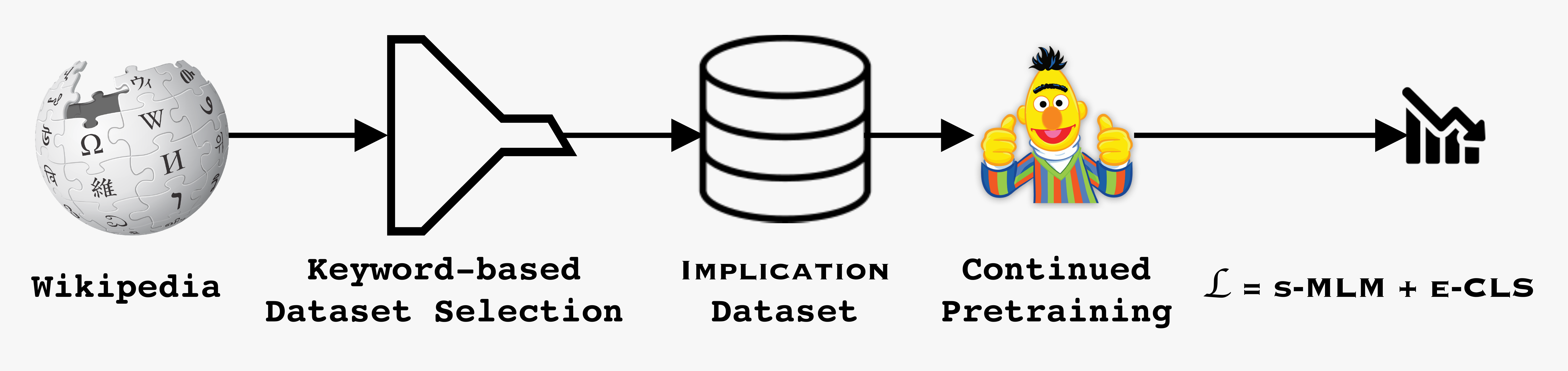}
	\caption{\small \textbf{Overview of \method{}}. We filter Wikipedia using specific logical keywords to create the \dataset{} dataset. This is then used for continued pretraining of a model using two loss objectives: selective masked language modeling (\selective{}) loss and entailment classification (\classification{}) loss. Please refer to Section \ref{sec:method} and Figure \ref{fig:loss_functions} for more details on the data selection process and loss function designs.}
	\label{fig:pipeline}
		\vspace{-0.3cm}
\end{figure*}

There have been some recent works on improving the logical reasoning abilities of PLMs \citep{lrreasoner,focalreasoner,merit}. These works typically generate a dataset containing symbolic structures such as logical graphs from text, logical contrast sets, etc., and then train the LM using custom loss objectives to learn logical reasoning abilities.
While the performance improvements achieved by these methods are encouraging, the proposed solutions generally require complex data processing to generate the additional structural information (graphs, contrast data, etc.) required for training the model. For example, \citet{merit} constructs synthetic context-answer pairs using the entity-level graph from Wikipedia for training the model. Further, the loss functions proposed in these works are very specifically designed in accordance with their respective data augmentation technique and widely differs from the typical masked language modeling loss used for LM pretraining \citep{devlin2019bert}. Additionally, some of these works usually require task-specific design choices, which are not necessarily learning generalizable logical reasoning ability that is reusable across different task formats. For example, \citet{lrreasoner} parses symbolic logical structures from the training data of a specific dataset, which might not generalize to a new dataset or task. 
Overall, it is unclear if these highly specific inductive biases are indeed essential for improving the logical reasoning abilities in language models, or if a simpler approach is possible.

On the other hand, prior works \citep{gururangan-etal-2020-dont} have shown that continual domain-adaptive pretraining of PLMs leads to performance gains on downstream tasks. Inspired by this, we propose \method{}, a continual pretraining-based approach to inject logical reasoning abilities in language models that requires minimal data processing and loss function modifications.

Firstly, we present a simple way of selecting sentences for training a model that is more likely to involve logical implications. We achieve this by defining a set of logical inference keywords and selecting a subset of sentences from a large text corpus, each containing at least one of these keywords. We hypothesize that PLMs can learn logical reasoning capabilities more easily using such sentences since the premise/conclusions are explicitly stated.
We note that in contrast to previous works \citep{gururangan-etal-2020-dont}, our method can select sentences from any general text corpus, eliminating the need for any domain-specific corpus.

Secondly, we modify the masked language modeling (MLM) loss \cite{devlin2019bert} to selectively mask specific words in the sentence, based on their parts-of-speech tags. Prior works \cite{lad2022using} have shown the benefit of selective masking of words on task-guided fine-tuning. We hypothesize that masking words with parts-of-speech (POS) tags that are related to higher-order reasoning (such as adverbs, conjunctions, etc.) present more challenging masked positions for the PLM to predict.
For instance, in Figure \ref{fig:example}, we observe that the words marked in blue boxes are more related to reasoning compared to the non-highlighted words that mainly involve knowledge about specific nouns or English grammar.

Lastly, we design a sentence-level classification loss to predict if the reasoning in the sentence describes an entailment in the reasoning process or a contradiction. 
This enables the model to better understand the differences between positive and negative implications in a sentence, thus improving logical reasoning.

To test \method{}, we evaluate it on two downstream logical reasoning tasks: \dataseta{} \cite{reclor_dataset} and \datasetb{} \cite{logiqa_dataset}, and compare it with other baselines. We achieve state-of-the-art performance on \datasetb{} and comparable performance on \dataseta{}. We demonstrate that our method generalizes across different model types. Further, we show that using our proposed loss functions does not induce any catastrophic forgetting \cite{catastrophic_forgetting} of the original language modeling skills. This demonstrates that our simple, continual pretraining approach is generalizable to different datasets and enables the PLM to acquire strong logical reasoning abilities.

\begin{figure*}[t]
	\centering
	\includegraphics[width=0.9\textwidth]{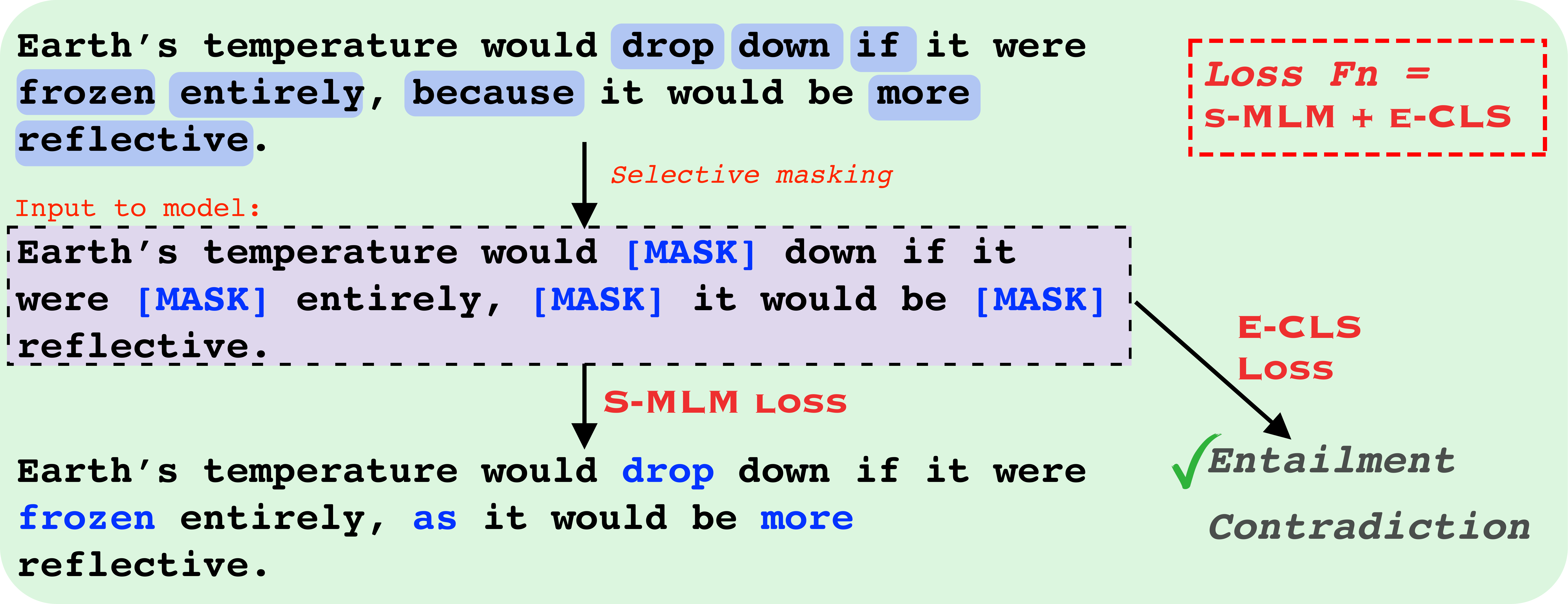}
	\caption{\small \textbf{Learning objectives in \method{}.} The selective masking step masks out words from a specific set of POS tags (the candidate words are shown in blue boxes). The \selective{} loss then predicts these masked words. The \classification{} loss classifies the masked sentence into one of two categories: entailment or contradiction. The overall loss function used in \method{} is the sum of both these objectives.}
	\label{fig:loss_functions}
		\vspace{-0.3cm}
\end{figure*}

Overall, compared to prior works, our proposed pretraining paradigm for \method{} 1) Uses sentences from text corpus for training instead of complex data structures such as entity graphs, etc. 2) Uses simple learning objectives that are closer to language modeling compared to the contrastive loss.  3) Is agnostic to both task format and downstream datasets. 4) Achieves state-of-the-art performance on \datasetb{}.


\section{Method}
\label{sec:method}
In this section, we describe the details of our proposed approach. In \method{}, we use a keyword-based selection strategy to collect a dataset of reasoning-related sentences called \dataset{} (\cref{sec:dataset_selection}) and then continue training a pretrained model checkpoint jointly using two loss functions (\cref{sec:loss_function}). This model is then fine-tuned on the training dataset of each task separately for evaluation. A detailed overview of the pipeline is shown in Figure \ref{fig:pipeline}.

\subsection{Dataset Selection}
\label{sec:dataset_selection}
PLMs are typically trained on web data which helps them to learn general language modeling capability. Then, PLMs are finetuned on downstream datasets to specialize on target tasks \citep{devlin2019bert,radford2018improving,raffel2019exploring}. Here, instead of focusing on a specific task, we want to teach the PLM generalizable logical reasoning abilities. We hypothesize that using training data that contains more logical sentences, rather than generic internet data, should help in improving the reasoning ability of the PLM. 

Although creating such a dataset automatically is a challenging task by itself, in \method{}, we explore a simple and intuitive way to create such a dataset. First, we select specific keywords that are typically encountered in sentences with logical implications. Broadly, we categorize these keywords into two types\footnote{Appendix \ref{app:keywords} presents the comprehensive list of keywords used to build the \dataset{} dataset.}:
\begin{itemize}
	\item \textbf{Positive implication (Entailment)}: These keywords are present in sentences where the reason generally entails the inference. Examples of such keywords would be ``therefore'', ``accordingly'', etc.
	\item \textbf{Negative implication (Contradiction)}: The keywords in this category are usually present in sentences where the reason contradicts the inference. For example, keywords such as ``but'', ``although'', etc., come under this category.
\end{itemize}

Next, we select sentences from Wikipedia such that they contain at least one of the keywords. We name this filtered version of Wikipedia as the \dataset{} dataset. While this keyword-based filtering does not necessarily ensure that the sentence has a logical implication, the retained data contains a higher portion of logically rich sentences than the general data. We argue that pretraining on this data helps the PLM to improve logical reasoning skills. Please refer to Appendix \ref{app:keywords} for more details on the list of keywords used to build the \dataset{} dataset.

\subsection{Learning objectives}
\label{sec:loss_function}

\paragraph{Selective masked language modeling (\selective{})} is a modified version of the masked language modeling (MLM) loss used in BERT \citep{devlin2019bert}. In the MLM loss, tokens in a sentence are masked at random and the model learns to predict the masked tokens. While this helps in learning a good language model, not all masked tokens require a similar degree of reasoning to predict them. In the example shown in Figure \ref{fig:loss_functions}, words such as ``were'', ``the'', etc. are decided more by the structure of the English language than any form of reasoning. In contrast, predicting logical words such as ``more'', ``and'' and ``hence'' would require more logical reasoning. 
Thus, we hypothesize that masking these logical words would likely teach the model to perform reasoning more effectively than masking a word at random.

While finding these exact logical words for a given sentence is a hard problem, in \method{} we simplify this by using a heuristic approach to consider words that belong to a specific set of parts-of-speech (POS) tags. More concretely, in \selective{} loss, we only randomly mask words with these 7 SpaCy POS tags \citep{spacy2}: ADJ, ADV, CONJ, CCONJ, PART, SCONJ, and VERB. Please refer to Section \ref{sec:ablation} for more empirical results that further justify this choice.

\paragraph{Entailment classification (\classification{})} Prior works have shown that semantic-aware sentence-level classification loss can be useful to learn the semantic information \cite{ernie_2.0}. Inspired by this, in addition to \selective{}, we use another auxiliary loss function that predicts whether a masked sentence contains some reasoning aspects that portray a sense of entailment or contradiction within the sentence. For example, in Figure \ref{fig:loss_functions}, the sentence is classified as ``Entailment'', because the phrase ``more reflective'' is entailed by the phrase ``frozen entirely''. We note that the input to the model is the same sentence with masked words that is used for \selective{} loss.
A model would ideally require strong logical reasoning abilities to understand the sentence and then predict if it refers to an entailment or contradiction. The labels for this loss are bootstrapped using the heuristic of checking the type of implication keyword present in the sentence (refer to Section \ref{sec:dataset_selection} for details). We note that although the keyword is a correlated feature that can be used to predict the label, on average the keyword would be masked out due to our selective masking policy, forcing the model to learn some logical semantics to minimize the loss. Additionally, even if the model predicts a wrong keyword in the sentence, it may still get the relationship between the sentences correctly. Therefore, the classification loss adds a stronger inductive bias specifically about the reasoning semantics in the sentence than the \selective{} loss.

\subsection{Continual Pretraining}
In \method{}, we combine both \selective{} and \classification{} objectives as a joint loss function to continually train a pretrained model checkpoint (Figure \ref{fig:pipeline}). Unlike prior works \citep{merit}, we don't need to add MLM loss to avoid catastrophic forgetting, as \selective{} is quite close to the standard MLM objective in format.

\subsection{Finetuning}
As our loss functions are task-format agnostic, we follow \citet{devlin2019bert} and add a randomly initialized MLP layer on top of the continually pretrained model. Then, we finetune the combined model on downstream datasets. 
\section{Experimental Setup}
\label{sec:setup}
In this section, we describe the details of the datasets on which we evaluate \method{}, the baselines we compare it with, and some implementation details of our training procedure.

\subsection{Datasets}
Following prior works \citep{merit}, we evaluate \method{} on two logical reasoning datasets:
\paragraph{\dataseta{} \citep{reclor_dataset}} is a reading comprehension dataset created from the logical reasoning questions from standardized graduate admission examinations. The test set is divided into two subsets: EASY (test-E) and HARD (test-H), where the EASY set contains instances whose options can be selected correctly without knowing the context and question. The train/dev/test split consists of 4,638/500/1,000 instances, respectively.
\paragraph{\datasetb{} \citep{logiqa_dataset}} is developed using publicly available logical examination papers for reading comprehension. The train/dev/test split consists of 7,376/651/651 instances, respectively.

\begin{table*}[t]
	\centering
	\resizebox{0.85\textwidth}{!}{%
		\begin{tabular}{m{4cm}m{1.2cm}m{1.2cm}m{1.2cm}m{1.2cm}m{1.2cm}m{1.2cm}}
			\toprule
			\multirow{2}{*}{\textbf{Model}}  & \multicolumn{4}{c}{\textbf{\dataseta{}}} & \multicolumn{2}{c}{\textbf{\datasetb{}}} \\
			\cmidrule(r){2-5} \cmidrule(r){6-7}
			& Dev & Test & Test-E & Test-H & Dev & Test \\
			\midrule
			RoBERTa			            & 62.6			        & 55.6 			& 75.5          & 40.0		& 35			        & 35.3 				\\
			DAGN			            & 65.2			        & 58.2 			& 76.1          & 44.1		& 35.5			        & 38.7 				\\
			\baselinea{}	            & 66.2			        & \textbf{62.4} 			& \textbf{81.4}          & \textbf{47.5}		& 38.1			        & 40.6 				\\
			\baselineb{}	            & 66.8	        & 58.9 			& 77.1          & 44.6		& 41.0			        & 40.3 				\\
			\baselinec{}	        & \textbf{67.8}                  & 60.7          & 79.6          & 45.9      & \textbf{42.4}                  & 41.5              \\
			\midrule
			\method{}		            & 67.2	& 58.2 			& 76.8          & 43.6		& 41.6	& \textbf{42.1}		\\
			\bottomrule
		\end{tabular}
	}
	\caption{Comparison of \method{} with other baselines on \dataseta{} and \datasetb{}. All the models are based on the RoBERTa-large model. The results for all the baselines are reported from \citet{merit}. Please refer to Section \ref{sec:overall_results} for more details.}
	\label{tab:main_results}
\end{table*}

\begin{table*}[t]
	\centering
	\resizebox{0.8\textwidth}{!}{%
		\begin{tabular}{lcccccc}
			\toprule
			\multirow{2}{*}{\textbf{Model}}  & \multicolumn{4}{c}{\textbf{\dataseta{}}} & \multicolumn{2}{c}{\textbf{\datasetb{}}} \\
			\cmidrule(r){2-5} \cmidrule(r){6-7}
			& Dev & Test & Test-E & Test-H & Dev & Test \\
			\midrule
			DeBERTa-v3                                      & 75.4 & 71.0 & 80.2 & 64.0 & 45.2 & 40.1 \\
			\method{} (DeBERTa-v3)                          & \textbf{76.8} & \textbf{72.8} & \textbf{81.8} & \textbf{65.7} & \textbf{48.4} & \textbf{44.4} \\
			\midrule
			DeBERTa-v2-xxlarge                              & 78.3 & 75.3 & 84.0 & 68.4 & 45.9 & 49.8 \\
			\baselinec{} (DeBERTa-v2-xxlarge)               & 80.6 & \textbf{78.1} & 84.6 & \textbf{72.9} & - & - \\
			\method{} (DeBERTa-v2-xxlarge)                  & \textbf{81.8} & 76.5 & \textbf{85.2} & 69.6 & \textbf{49.6} & \textbf{51.0} \\
			\bottomrule
		\end{tabular}
	}
	\caption{Comparison of \method{} with other baselines on \dataseta{} and \datasetb{} with DeBERTa as the base architecture. Results for \baselinec{} are reported from \citet{merit}, which is missing results on \datasetb{}. Other baselines are reproduced by ourselves. The base models are shown in brackets. Please refer to Section \ref{sec:overall_results} for more details.}
	\label{tab:results_deberta}
\end{table*}

\subsection{Baselines}
We compare the accuracy of \method{} with the following baselines: \baselinea{} \citep{lrreasoner}, DAGN \cite{huang-etal-2021-dagn}, \baselineb{} \citep{focalreasoner}, and \baselinec{} \citep{merit}.

\subsection{Implementation Details}
For creating the \dataset{} dataset, we use the Wikipedia version provided under HuggingFace Datasets \citep{wolf-2020HuggingFace-datasets} as the main corpus.\footnote{\url{https://huggingface.co/datasets/wikipedia}} The list of keywords we use for filtering sentences from Wikipedia are listed in Appendix \ref{app:keywords}. We experiment with RoBERTa-Large \citep{liu2019roberta}, DeBERTa-v3 \citep{he2021debertav3}, and DeBERTa-v2-xxlarge \citep{he2020deberta} as the base models for \method{}. We pretrain the last two layers of the Transformer \citep{vaswani2017attention} layer for 3 epochs, using a batch size of 4096. Please refer to Appendix \ref{app:hyperparams} for more details on training and finetuning hyperparameters.
\section{Results}

\subsection{Overall Results}
\label{sec:overall_results}
In this section, we compare the performance of \method{} with prior baselines on the two logical reasoning datasets for different base architectures. The results of using pretrained Roberta-Large as the starting checkpoint for our method are shown in Table \ref{tab:main_results}. We observe that \method{} outperforms all baselines on \datasetb{} and performs lower on \dataseta{} than three baselines, although consistently outperforming the RoBERTa baseline. Overall, this demonstrates that our simple continual pretraining approach is indeed strong enough to perform well on logical reasoning tasks as compared to the prior models that depend on much more complex training data and loss function designs.

\begin{table*}[t]
	\centering
	\resizebox{0.8\textwidth}{!}{%
		\begin{tabular}{lcccccccccc}
			\toprule
			\multirow{1}{*}{\textbf{Model}} & MNLI & QNLI & QQP & RTE & SST & MRPC & CoLA & STS && \textbf{Avg}\\
			\midrule
			RoBERTa-Large & 90.2 & 94.7 & 92.2 & 86.6 & 96.4 & 90.9 & 68.0 & 92.4 && 88.9 \\
			\method{} & 90.3 & 94.9 & 92.1 & 88.1 & 96.2 & 92.2 & 68.6 & 91.9 && \textbf{89.3} \\
			\bottomrule
		\end{tabular}
	}
	\caption{Performance on the dev set of GLUE benchmark. Following \citet{devlin2019bert}, we do not report performance on the WNLI dataset. Please refer to Section \ref{sec:glue_results} for further details.}
	\label{tab:glue_results}
\end{table*}

To test the generality of our approach across different architectures, we use pretrained DeBERTa-v3 and DeBERTa-v2-xxlarge as the base models for continued training. The results of using these models are shown in Table \ref{tab:results_deberta}. We find that \method{} outperforms both the baselines on both datasets. Further, we observe that \method{} performs 1.5\% worse compared to \baselinec{} on \dataseta{} test set. This shows that our continual pretraining process can improve performance across different LM architectures.

\subsection{Performance on GLUE Benchmark}
\label{sec:glue_results}
While improving the logical reasoning abilities of a PLM is important, it is equally important to retain the natural language understanding skills learned during pretraining. To demonstrate that our proposed approach does not lead to catastrophic forgetting, we finetune \method{} on each dataset of the GLUE benchmark \cite{wang2019glue} and evaluate the finetuned checkpoint on the Dev set. The results are compared with the Dev set results for the RoBERTa model \cite{roberta} in Table \ref{tab:glue_results}. Following \citet{devlin2019bert}, we omit the evaluation on the problematic WNLI set. Overall, we observe that \method{} can slightly improve the overall performance on the GLUE benchmark.
This demonstrates that our proposed continued pretraining strategy is able to learn better logical reasoning abilities without any catastrophic forgetting of general-purpose language modeling skills, and these logical reasoning capabilities are also beneficial for general natural language understanding.

\subsection{Qualitative Analysis}
\label{sec:qualitative}
\begin{figure}[t]
	\centering
	\includegraphics[width=0.9\columnwidth]{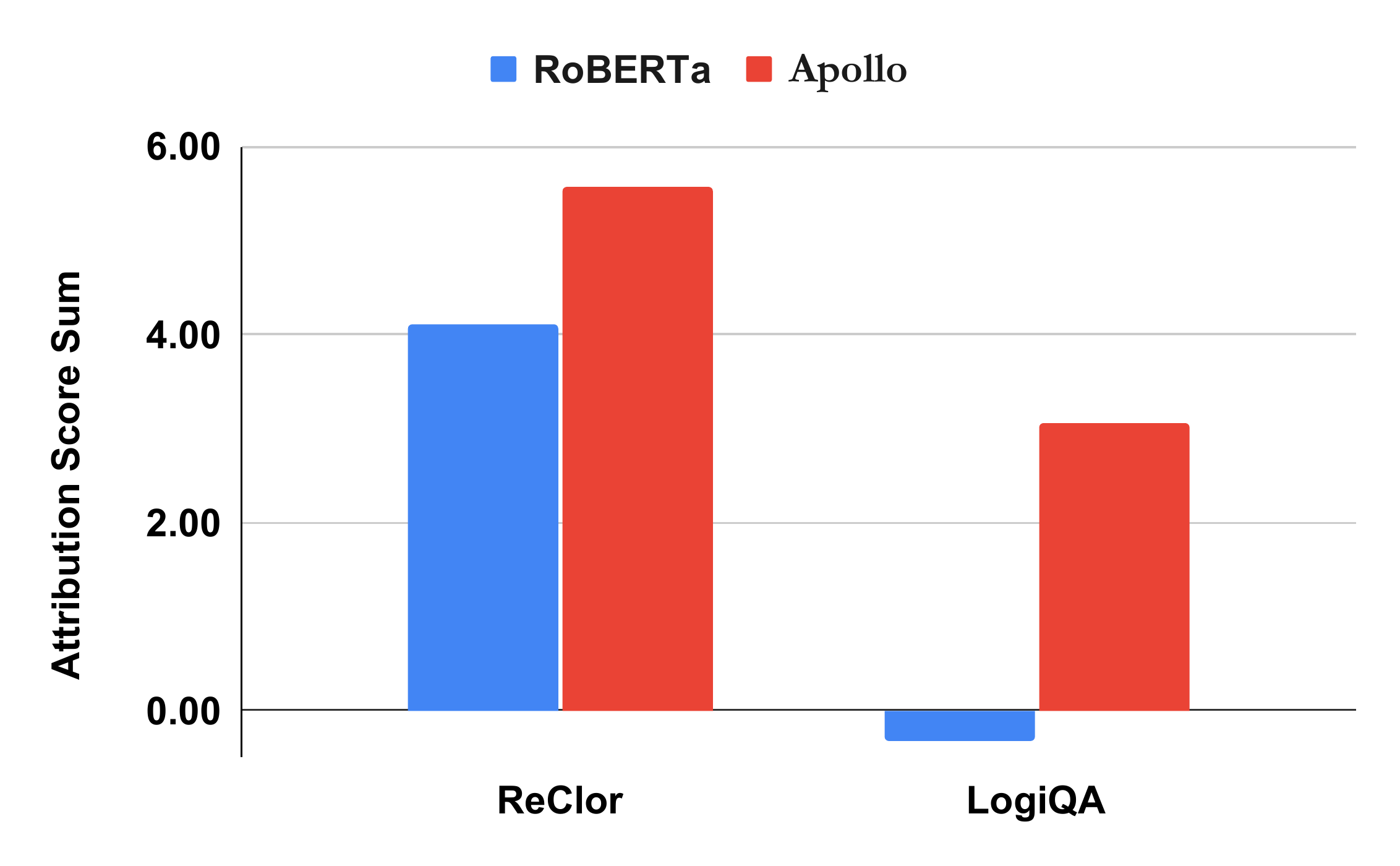}
	\caption{\small Comparison plot of the keyword attribution scores between RoBERTa-large and \method{}. Please refer to \ref{sec:qualitative} for more details.}
	\label{fig:attribution_plot}
		\vspace{-0.3cm}
\end{figure}

In this section, we analyze the effect of continued pretraining on the model's overall faithfulness. Post-hoc interpretability methods such as Integrated Gradients \cite{sundararajan2017axiomatic}, are algorithms to determine the importance of words in the input towards predicting a particular class. These importance scores are also referred to as \textit{attribution scores}. To approximate the impact of continued pretraining, we compute the overall change in attribution scores for the implication keywords, before and after pretraining the model using our proposed datasets and loss functions. Specifically, we compute the sum of the attribution scores for the keywords present in each instance of the validation set. The results are shown in Figure \ref{fig:attribution_plot}. We observe that our proposed pretraining increases the overall attribution score by a significant margin, indicating that the model intrinsically learns these important logical keywords, which is desirable.

\subsection{Ablation Studies}
\label{sec:ablation}
In this section, we ablate various design choices in constructing the \dataset{} dataset, and our proposed method. For the ablations involving \method{}, we use RoBERTa-Large as the base model and the \dataset{} dataset, if not mentioned separately. All the reported numbers are on the validation set of the downstream task, since we used these ablation studies in our model's design choices.

\begin{table}[t]
	\centering
        \resizebox{\columnwidth}{!}{%
		\begin{tabular}{lcc}
			\toprule
			\multirow{1}{*}{\textbf{Model (Dataset, Loss functions)}}  & \multicolumn{1}{c}{\textbf{\dataseta{}}} & \multicolumn{1}{c}{\textbf{\datasetb{}}} \\
			\midrule
			RoBERTa (\random{}, MLM)                                 & 60.2 & 35.0 \\
			RoBERTa (\random{}, \selective{})                        & \textbf{63.8} & \textbf{36.4} \\
			\midrule
			RoBERTa (\dataset{}, MLM)                                & 64.8 & 36.6 \\
			RoBERTa (\dataset{}, \selective{})                       & 65.4 & 41.5 \\
			RoBERTa (\dataset{}, \selective{} + \classification{})   & \textbf{67.2} & \textbf{41.6} \\
			\bottomrule
		\end{tabular}
  }
    \vspace{-0.1cm}
	\caption{Effect of \dataset{} dataset and the loss functions on the dev set performance of \dataseta{} and \datasetb{}.}
		\vspace{-0.2cm}
	\label{tab:ablation_data_loss}
 \vspace{-0.2cm}
\end{table}

\begin{table}[t]
	\centering
	\resizebox{0.85\columnwidth}{!}{%
		\begin{tabular}{lcc}
			\toprule
			& \multicolumn{1}{c}{\textbf{\dataseta{}}} & \multicolumn{1}{c}{\textbf{\datasetb{}}}\\
			\midrule
			\textbf{Keyword Category} \\
			\midrule
			\dataset{}-Positive  & 65.0 &	38.6 \\
			\dataset{}-Negative  & 64.6 &	37.6 \\
			\dataset{}                  & \textbf{65.4} &	\textbf{41.5} \\
			\midrule
			\textbf{POS Category} \\
			\midrule
			Base  					& \textbf{65.4} &	\textbf{41.5} \\
			Base + Nouns 			& 64.0 &	39.0 \\
			Base + Nouns + Random 			& 64.8 &	36.6 \\
			\bottomrule
		\end{tabular}
	}
	\caption{Ablation of design choices involved in keyword-based dataset selection and \selective{} loss function implementation. We report the performance on the dev set of each dataset. Please refer to Section \ref{sec:ablation} for more details.}
	\label{tab:dataset_ablation}
\end{table}

\paragraph{Effect of datasets and loss functions} To study the effect of using \dataset{} for continued pretraining along with the proposed loss functions, we first create \random{}, a random subset of Wikipedia of similar size as that of \dataset{}, and also consider using the standard masked language modeling (MLM) loss \cite{devlin2019bert}, where any token can be masked at random. The results of the ablation are shown in Table \ref{tab:ablation_data_loss}. We observe that using the \dataset{} dataset leads to consistent improvements on both datasets when compared to the \random{} dataset. Additionally, we find that both the \selective{} and \classification{} loss lead to improvements over MLM loss. Thus, this empirically justifies our choice of the dataset and loss functions proposed here.

\paragraph{Effect of keyword category} In this ablation, we study the effect of the keyword categories that we use for filtering Wikipedia. For this, we create two different pretraining datasets \dataset{}-Positive and \dataset{}-Negative using the positive and negative implication keywords, respectively (refer to Section \ref{sec:dataset_selection}). The total number of sentences in these datasets is 7.5M and 11.3M, respectively. Our complete dataset \dataset{} thus has a total of 18.3M sentences. The results of the ablation are shown in Table \ref{tab:dataset_ablation}, under the section ``Keyword Category''. We observe that \dataset{}-Positive, although smaller in size, leads to better performance on both downstream tasks, compared to \dataset{}-Negative. One reason for this is that the sentences with positive keywords are more likely related to reasoning than the negative counterparts because the negative keywords are used in many diverse scenarios in the English language. For example, the word ``\textit{still}'' can be used in a non-logical manner such as ``\textit{I am still waiting for the call}''. Overall, we observe that the combined \dataset{} dataset leads to the best performance, demonstrating that both the positive and negative implication keywords are essential to improve logical reasoning.

\paragraph{Effect of POS tag category} In this, we analyze the effect of the parts-of-speech (POS) tags we use to mask tokens in our \selective{} loss. We consider the following categories:
\begin{itemize}[noitemsep,nolistsep]
	\item \textbf{Base}: This consists of the POS tags used in \method{}, i.e., ADJ, ADV, CONJ, CCONJ, PART, SCONJ, and VERB.
	\item \textbf{Nouns}: Here, we consider the tags referring to nouns and pronouns, i.e., NOUN, PRON, and PROPN.
	\item \textbf{Random}: This consists of remaining categories such as ADP, INTJ, DET, PUNCT, etc.
\end{itemize}
To study the effect of the POS tags, we incrementally add the ``Nouns'' and ``Random'' categories to the base case and evaluate the effect of pretraining using the \selective{} loss. The results of this ablation are shown in Table \ref{tab:dataset_ablation}, under the section ``POS Category''. We observe that masking nouns and pronouns (``Nouns'') leads to a significant performance drop. We attribute this drop to the fact that predicting a correct noun in a sentence would likely require more world knowledge than logical reasoning. Using the remaining categories for selective masking (``Random''), effectively making the loss function equivalent to random MLM, leads to some drop in performance as well, indicating that our set of POS tag categories is indeed more useful to learn logical reasoning.

\begin{figure}[t]
	\centering
	\includegraphics[width=\columnwidth]{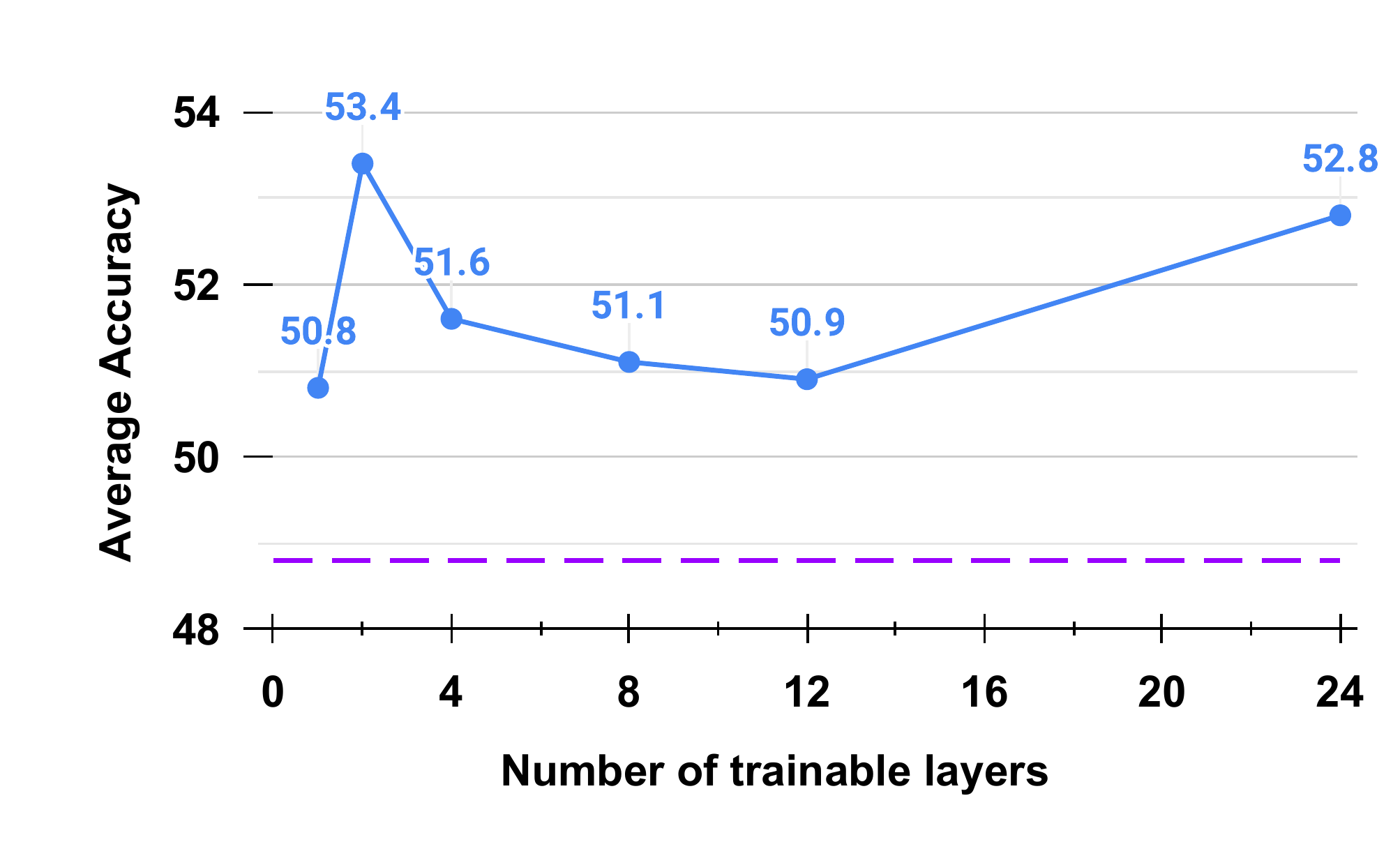}
	\caption{\small Average performance on the dev set of \dataseta{} and \datasetb{} with increasing number of trainable layers of \method{}. The pink dashed line shows the average performance of RoBERTa-Large when all layers are finetuned. Please refer to Section \ref{sec:ablation} for more details.}
	\label{fig:ablation_layer}
\end{figure}

\paragraph{Effect of the number of trainable layers} In order to study the effect of training different numbers of parameters of the RoBERTa model, we vary the number of trainable layers of the transformer architecture between $1$ and $24$ (i.e., training the complete model). The results are shown in Figure \ref{fig:ablation_layer}. The blue solid line shows the performance of \method{} and the purple dashed line denotes the average performance of RoBERTa-Large when all layers are finetuned. From the plot, we observe that with increasing the number of trainable layers, the performance improves till layer $2$, and then continues to degrade until all the layers are being trained. Prior works \cite{tenney-etal-2019-bert} have shown that PLMs learn syntactic-level information in the lower layers of the transformer and semantic-level information in the upper layers. Thus, we hypothesize that the logical reasoning task initially benefits from an increasing number of trainable layers, as the semantic information needed to understand logic is being captured. But lower layers that contain the syntactic information do not benefit as much when trained using the same data as they are less related to high-level logical reasoning. The full model finetuning surprisingly performs quite well as all the model layers along with the token embeddings are being trained specifically for the logical reasoning task. But it takes significantly larger compute to finetune such a model. Overall, we find that by training the topmost two layers of the model, we are able to achieve the best performance on both datasets and hence we follow this across all variants of \method{}.
\section{Related Works}

\paragraph{Logical Reasoning in LMs} Reasoning in natural language has been a prevalent problem in NLP. In recent years, logical reasoning in textual data has seen an increasing focus. \dataseta{} \citep{reclor_dataset} and \datasetb{} \citep{logiqa_dataset} are reading comprehension-style datasets focused on questions that require reasoning using information from a given context. Prior works have predominantly used language models \cite{lrreasoner,merit} or graph neural networks (GNNs) \cite{huang-etal-2021-dagn,logiformer,li-etal-2022-adalogn,focalreasoner} to perform logical reasoning over text. \citet{lrreasoner} proposed \baselinea{}, which parses symbolic logical structures from the training data of \dataseta{} for data augmentation using logical context extensions to train a PLM. \citet{merit} proposed \baselinec{}, that used Wikipedia to generate sentence pairs for contrastive learning that are logically related, and trained the PLM using contrastive loss. DAGN \cite{huang-etal-2021-dagn} uses the discourse structure of the texts to perform logical reasoning using GNNs. \baselineb{} \cite{focalreasoner} constructs logical graphs using the chain of facts present in a task instance and uses GNNs to reason on the graph. GNN-based methods are not directly in scope since our main objective is to improve the logical reasoning skills of language models. Following \cite{merit}, we compare our method with two GNN-based representative methods DAGN and \baselineb{}.

Both \baselinea{} and \baselineb{} use data augmentation that is specific to the task being solved, making the pretraining process specific to the downstream dataset, and thus not generalizable across tasks. While \baselinec{} addresses this issue by using Wikipedia to generate logical graphs, their contrastive loss formulation requires counterfactual data augmentation, which potentially distorts the factual knowledge present in the pretrained model. Additionally, their approach is restricted to using Wikipedia as the data source since they heavily rely on forming entity graphs from Wikipedia texts. In contrast, we propose a simple continued pretraining strategy by modifying the masked language modeling loss \citep{devlin2019bert} and sentence classification loss to improve the logical reasoning ability of language models. Our approach is simple to integrate during pretraining, is not dependent on any data processing, and generalizes well across different datasets.

Along a related line, \citet{ruletaker} used synthetically generated data to teach PLMs to perform logical deductions over a given rule base to predict the entailment of a hypothesis. This led to some recent developments in trying to build systems that can generate step-by-step reasoning chains that demonstrate the model's reasoning process \citep{prover,proofwriter,fairr}. While this progress is encouraging, the use of synthetic data for training the models limits the generality of the logical reasoning skills learned by these models. Recent works have questioned if these models are indeed learning to perform logical reasoning in a robust manner or just learning some shortcuts from training data \citep{sampling_algorithm,robustlr}. In contrast, our method uses real-world sentences which alleviates the issue of using synthetic datasets for reasoning.

\paragraph{Selective masking} A key step in the processing of masked language modeling loss \citep{devlin2019bert} is to determine which tokens to mask. Originally, \citet{devlin2019bert} \textit{randomly} mask 15\% of tokens. Prior works have tried different techniques to select which tokens to mask. For example, ERNIE \citep{zhang-etal-2019-ernie} and EntityBERT \citep{entitybert} mask named entities to perform better knowledge-driven tasks. Other prior works \citep{gu-etal-2020-train,lad2022using} calculate the importance of words for a specific task and selectively mask the most important words. In this work, we explore the use of selective masking in the context of logical reasoning, using a novel heuristic of selecting specific POS-tagged words.
\section{Conclusion}
In this paper, we proposed \method{}, an adaptive pre-trained language model with logical reasoning abilities. We use a subset of Wikipedia sentences for continued pretraining of the model using two self-supervised loss functions. The choice of the training dataset and loss functions are guided by the goal to include more reasoning-related sentences and training signals, respectively. Through experiments on two logical reasoning datasets and ablation studies, we demonstrate the effectiveness of our proposed approach. Overall, we show that \method{} is a generalized solution to improving logical reasoning in language models.

A key advantage of \method{} is that the pretraining steps are independent of the dataset used to train the model and the downstream task format. This opens the scope to use a larger text corpus for training such as C4 \cite{raffel2019exploring}. Additionally, expanding on the keywords beyond positive and negative implications (for example, conditionals such as ``if-then'', ``either-or'', etc.) can also benefit the training pipeline.

\section{Limitation}
A limitation of this approach is the trade-off between completeness and noise in the training data. 
While our method using keywords to extract text from Wikipedia is effective, \dataset{} likely contains redundant sentences that cannot improve the model's logical reasoning capability. A better rule-based or neural model might be able to extract a better corpus with potentially higher computational costs.
Additionally, using POS tagging limits the application of this approach to languages with well-defined POS taggers. Switching to a more universal semantic tagging system \cite{abzianidze-bos-2017-towards} can potentially alleviate this.

\section*{Acknowledgements}
This research is supported in part by the Office of the Director of National Intelligence (ODNI), Intelligence Advanced Research Projects Activity(IARPA), via Contract No. 2019-19051600007, the DARPA MCS program under Contract No.N660011924033, the Defense Advanced Research Projects Agency with award W911NF-19-20271, NSF IIS 2048211, NSF SMA 1829268, and gift awards from Google, Amazon, JP Morgan, and Sony. We would like to thank all the collaborators in the USC INK research lab for their constructive feedback on the work.

\bibliography{custom}

\clearpage

\appendix
\section{List of Keywords}
\label{app:keywords}
In this section, we list the set of keywords that we use to filter the entire WikiPedia data. Any sentence that contains one of the keywords is considered as part of our filtered dataset \dataset{}. The keywords are divided into two types as described below:
\begin{itemize}
	\item \textbf{Positive implication (Entailment)}: These keywords are present in sentences where the reason generally entails the inference. Examples of such keywords would be ``therefore'', ``accordingly'', etc. We consider the following keywords in this category: ``therefore'', ``accordingly'', ``so'', ``thus'', ``consequently'', ``hence'', ``thence'', ``and so'', ``for this reason'', ``in consequence'', ``on account of'', ``on the ''grounds'', ``since'', ``therefrom'', ``thereupon'', ``to that end'', ``whence'', and ``wherefore''.
	\item \textbf{Negative implication (Contradiction)}: The keywords in this category are usually present in sentences where the reason contradicts the inference. For example, keywords such as ``but'', ``although'', etc., come under this category. Here, we consider the following keywords: ``but'', ``although'', ``however'', ``nevertheless'', ``on the other hand'', ``still'', ``though'', and ``yet''.
\end{itemize}

\section{Hyperparameter Details}
\label{app:hyperparams}
In continual pretraining, we select the learning rate from the set $\{7e-6, 1e-5, 7e-5\}$, batch size $4$, gradient accumulation step size from the set $\{64, 128\}$, warmup ratio $0.1$, and train the model on a cluster of $8$ A100 GPUs. To fine-tune a continually pretrained checkpoint, we use the training data of each dataset separately. We select learning rate from the set $\{8e-6, 1e-5, 5e-5\}$, batch size of $4$, and gradient accumulation step size $1$. To train the models we use a cluster of $8$ A100 GPUs, which typically takes around 20 hours for the largest model.

\end{document}